\documentclass[english]{article}

\usepackage[affil-it]{authblk}
\usepackage{graphicx}
\usepackage[space]{grffile}
\usepackage{latexsym}
\usepackage{textcomp}
\usepackage{longtable}
\usepackage{multirow,booktabs}
\usepackage{amsfonts,amsmath,amssymb}
\usepackage{url}
\usepackage[utf8]{inputenc}
\usepackage{hyperref}
\hypersetup{colorlinks=false,pdfborder={0 0 0}}
\usepackage[table]{xcolor}% http://ctan.org/pkg/xcolor
\usepackage{multirow}
\usepackage{array}
\usepackage{listings}
\usepackage{lineno}
\usepackage{color}
\usepackage{babel}
\usepackage{csquotes}
\usepackage{ltablex}
\usepackage{soul}

\usepackage{amsmath}
\usepackage{amssymb}
\usepackage{amsthm}

\MakeOuterQuote{"}
\definecolor{mygreen}{rgb}{0,0.2,0}
\definecolor{mygray}{rgb}{0.95,0.95,0.95}
\definecolor{mymauve}{rgb}{0.58,0,0.82}

\definecolor{lbcolor}{rgb}{0.95,0.95,0.95}

\newtheorem{theorem}{Theorem}
\newtheorem*{theorem*}{Theorem}

\begin{document}

\lstset{
  basicstyle=\ttfamily,
  columns=fullflexible,
  keepspaces=true,
  breaklines=true,
  backgroundcolor=\color{mygray},
  keywordstyle=\color{blue},
  stringstyle=\color{mymauve},
  ndkeywordstyle=\color{red},
  commentstyle=\color{mymauve},
  identifierstyle=\color{mygreen}
}

\lstdefinelanguage{giac}{
  keywords={factor, eliminate},
  ndkeywords={>>},
  sensitive=true
}
\lstdefinelanguage{mylog}{
  comment=[l]{//},
  morecomment=[s]{/*}{*/}
}

%\title{Enhancements \\in Geometry Theorem Proving \\in GeoGebra}
\title{GeoGebra Tools with Proof Capabilities}

\author{Zolt{\'{a}}n Kov{\'{a}}cs}
\affil{Private University College of Education \\of the Diocese of Linz, Austria}

\author{Csilla S\'olyom-Gecse}
\affil{Babe\c{s}-Bolyai University, Cluj-Napoca, Romania}

\date{\today}

\bibliographystyle{apalike}

\maketitle

\tolerance10000

\begin{abstract}
We report about significant enhancements of the complex algebraic geometry theorem proving subsystem in GeoGebra for automated proofs in Euclidean geometry, concerning the extension of numerous GeoGebra tools with proof capabilities. As a result, a number of elementary theorems can be proven by using GeoGebra's intuitive user interface on various computer architectures including native Java and web based systems with JavaScript. We also provide a test suite for benchmarking our results with 200 test cases.
\end{abstract}

\section{Introduction}

GeoGebra is an educational mathematics software tool, with millions of users. In 2005, its founder Markus Hohenwarter broadened its software development into
an open source project. GeoGebra's features (including dynamic geometry, computer algebra, spreadsheets and function investigation) primarily focus on facilitating student experiments in Euclidean geometry, and not on formal reasoning. Including automated deduction tools in GeoGebra's dynamic geometry system (DGS) could
introduce a whole new range of learning and teaching scenarios.

Since automated theorem proving (ATP) in geometry has reached a rather mature stage, in 2010 some ATP
experts agreed on starting a project of incorporating and testing a number of different automated provers for geometry in GeoGebra. This collaboration was
initiated by Tom\'as Recio. Since the initial kickstart this project reached the following milestones:
\begin{enumerate}
\item A workshop for {\bf theoretical planning} took place in Santiago de Compostela, Spain, February 2011.
\item A second workshop for {\bf implementation planning} took place in Alcal\'a de Henares, Spain, January 2012.
\item A {\bf prototype} implementation was presented in Alcal\'a de Henares in June 2012 by demonstrating 44 test cases using 5 different theorem prover methods \cite{Botana_2012, Kovacs_2012}.
\item {\bf First public release} in GeoGebra 5.0 in October 2014 with 60 test cases \cite{Botana_2015}.
\item Full {\bf documentation} and fixing several issues according to users' feedback in July 2015 in \cite{Kovacs_2015b}.
\item {\bf Extension} of the set of the translated dynamic geometry construction tools to cover 200 test cases.
\end{enumerate}

In this paper we report about the last milestone. In section \ref{overview} we give a comprehensive overview about the first milestones. Section \ref{symeqs} summarizes our results by focusing on the general improvements in GeoGebra. Section \ref{benchmarks} shows some tables concerning our test results. Section \ref{future} sketches up our next planned steps for another milestone in the implementation.

\section{Overview}
\label{overview}

An interactive prover system designed mainly for secondary school students can differ from expert prover systems in some aspects. For example, {\it GCLC} \cite{Jani_i__2006} and {\it OpenGeoProver} \cite{OGP} process a program code written in its special language and print the output as a precise report about the computation details. By contrast, a DGS tool should collect all pieces of information about the relationships of the objects purely by analyzing the construction being created by point-and-click edits and possibly some other input parameters for the prover commands; finally the output is typically a yes/no answer and eventually some extra prescribed conditions to avoid degeneracy cases.

There is a plenty of literature on reports on successful applications of DGS by extending one with an ATP subsystem. Among them, here we mention \textit{GeoProof} \cite{Narboux2007b} and \textit{LADucation} \cite{Botana_2009} which are open-sourced, and thus it is possible to continue their efforts by external researchers also.
A publicly available variant of LADucation was already able to import a GeoGebra construction and set up an equation system which was solved by an external computer algebra system (CAS).
Other systems including \textit{JGEX} \cite{Ye_2011} and \textit{Cinderella} \cite{ulli99} are not open-sourced, but built upon a similar approach: visualizations in the DGS must be supported by ATP computations.

Our project harnessed GeoGebra's success in the classrooms and tried to address some problems of the existing DGS/ATP prototypes including small distribution, being unmaintained or incomplete operation. In our solution in GeoGebra a user creates a dynamic geometry construction which contains free points and dependent points as usual. All dependent points are already determined by the free points, however, all free points can be dragged by the user as desired. When a free point is dragged, some dependent points will also be changed by following the definitions in the construction steps. In such a way geometric theorems can be visualized by experiment.

This technique is well known in the world of DGS. Going one step forward, an ATP subsystem can give a more sound answer whether the visually obvious facts (for example, if three dependent points in a given construction are always collinear) generally hold. GeoGebra's command line interface with its {\bf Prove} and {\bf ProveDetails} commands and the graphical {\it Relation Tool} \cite{Kovacs_2015c} introduce a higher level interface to investigate the problem setting by using an ATP subsystem.

Proving Euclidean elementary geometry theorems was introduced in GeoGebra with its version 5 in September 2014.
A report \cite{Botana_2015} shows a benchmark about 60 theorems which can be directly checked with the {\bf Prove} and {\bf ProveDetails} commands in GeoGebra. More details are shown in \cite{Kovacs_2015a} about how the prover subsystem is embedded to GeoGebra's user interface intuitively by using and extending the Relation Tool.

There are several approaches to compute a proof internally by using GeoGebra's {\it portfolio prover} \cite{Kovacs_2014}, including
\begin{itemize}
\item Wu's method \cite{Wu_1984} by using OpenGeoProver externally, and also
\item the area method \cite{Chou_1993} (via OpenGeoProver), moreover
\item Recio's exact check method \cite{Kovacs_2012} and
\item the Gr\"obner basis method \cite{Kapur_1986,Kutzler_1986}.
\end{itemize}

In our present work we focused on the internally implemented \textit{Gr{\"{o}}bner basis method} which translates the geometric objects to algebraic equations directly and manipulates on the algebraic equation system by eliminating the dependent variables. Our work could be however used for Wu's method also, since we just defined a set of equations to translate geometric construction tools into an algebraic approach. We used complex algebraic geometry in our computations which is a standard way to set up a Euclidean geometry question (see \cite[chapter 6]{Cox_2007}).

\section{Our enhancements}
\label{milestone6}

We report about our contributions to GeoGebra in two major areas:
\begin{enumerate}
\item Implementation of symbolic equations for various geometric tools (section \ref{symeqs}).
\item Creating a number of tests to extend the benchmarks (section \ref{benchmarks}).
\end{enumerate}

\subsection{Symbolic equations}
\label{symeqs}

GeoGebra's geometry tools have been classified by \cite[p. 104]{Preiner_2008}  as ``easy to use'', ``middle'' and ``difficult to use''. Preiner defines
two criteria for a tool to be easy (p. 121):
\begin{enumerate}
\item
The tool does not depend on already existing objects, or just requires
existing points which can also be created `on the fly' by clicking on the drawing pad. The
order of actions is irrelevant and no additional keyboard input is required.
\item
The tool directly affects only one type of existing object or all
existing objects at the same time and requires just one action. Again, the order of actions
is irrelevant and no additional keyboard input is required.
\end{enumerate}

The basic concept in our work was to implement theorem proving features for the easier tools in GeoGebra. Also it was important
that the usually discussed classroom theorems can be quickly constructed by using the easier tools. The classroom theorems usually require points, segments, rays,
lines and circles, and angles.
For some more advanced topics tangents, parabolas, ellipses and hyperbolas may be needed.

\subsubsection{Translating geometry to algebra}

Implementing {\it angles} and {\it conics} may have theoretical difficulties in our approach. For {\it angles}, we refer to the fact that it is not possible to define only the interior bisector of an angle: we always need to work together with internal and external angles at the same time (cf.~\cite[p.~40]{Chou_1987}). This is a consequence of handling angles: there is no way to check equality unless one computes the tangent of them, that is, instead of checking $\alpha=\beta$ one verifies $\tan(\alpha)=\tan(\beta)$ and these formulas are equivalent only if we set up some restrictions, say $0\leq\alpha,\beta<\pi$. In this sense we cannot distinguish $0^{\rm o}$ and $180^{\rm o}$.

For {\it conics}, ellipses and hyperbolas must also be handled as non-distinguishable objects, because using the synthetic approach we need to define them with their foci, and the defining relations are the same. More precisely, given foci $A$ and $B$ and conic point $C$, another point $P$ is an element of the conic if and only if $AC+CB=AP+PB$ in the case of an ellipse and $|AC-CB|=|AP-PB|$ (that is, $(AC-CB)^2=(AP-PB)^2$) in the case of a hyperbola. Since the lengths in these equations are non-negative quantities, we either need to add constraints $AC\geq0, CB\geq0, AP\geq0$ and $PB\geq0$ (which are not possible in complex algebraic geometry due to lack of inequalites), or we need to use the squared quantities $AC^2$, $CB^2$, $AP^2$ and $PB^2$ and express these equations exclusively by them. In this second case we need to eliminate the non-squared quantities from the equation. With the help of the following computer algebra command we learn that for both the ellipse and the hyperbola we get the same product of 8th degree (here we used Giac \cite{Kovacs_2015a} for computations):
\begin{lstlisting}[language=giac]
>> factor(eliminate([AC+CB=AP+PB,AC^2=ac^2,CB^2=cb^2,AP^2=ap^2,PB^2=pb^2],[AC,CB,AP,PB]))
\end{lstlisting}

\noindent returns
\begin{lstlisting}[language=giac]
[(ac-cb-ap-pb)*(ac-cb-ap+pb)*(ac-cb+ap-pb)*(ac-cb+ap+pb)*(ac+cb-ap-pb)*(ac+cb-ap+pb)*(ac+cb+ap-pb)*(ac+cb+ap+pb)]
\end{lstlisting}

\noindent which has the same result as for the input

\begin{lstlisting}[language=giac]
>> factor(eliminate([(AC-CB)^2=(AP-PB)^2,AC^2=ac^2,CB^2=cb^2,AP^2=ap^2,PB^2=pb^2],[AC,CB,AP,PB]))
\end{lstlisting}

\noindent Interpreting the result, it is only possible to define the set $AC\pm CB=\pm AP\pm PB$ in the complex algebraic geometry sense which consists of 8 theoretical curves:
\begin{enumerate}
\item $AC+CB=AP+PB$, the ellipse,
\item $AC+CB=AP-PB\Leftrightarrow AC+CB+BP=AP$, which---according to the triangle inequality---is possible only in a degenerate case when $A$, $B$ and $C$ (and also $P$) are collinear,
\item $AC+CB=-AP+PB\Leftrightarrow PA+AC+CB=PB$, similar to the previous collinear case,
\item $AC+CB=-AP-PB\Leftrightarrow AC+CB+AP+PB=0$ which is possible only in a degenerate case when $A=B=C=P$,
\item $AC-CB=AP+PB\Leftrightarrow CA+AP+PB=CB$, similar to the former collinear cases,
\item $AC-CB=-AP+PB$, one branch of the hyperbola,
\item $AC-CB=AP-PB$, the other branch of the hyperbola,
\item $AC-CB=-AP-PB\Leftrightarrow CA+AP+PB=CB$, again similar to the former collinear cases.
\end{enumerate}
That is, we indeed obtained that an ellipse and a hyperbola cannot be distinguished in this model (but all other non-degenerate curves can be distinguished from them). This issue will give some limitations to investigate special features of conics, but still enable investigating some common features of them. For example, the following generalization of Pascal's hexagon theorem for conics holds (see Fig.~\ref{Pascal}):

\begin{figure}[ht!]
\includegraphics[width=\textwidth]{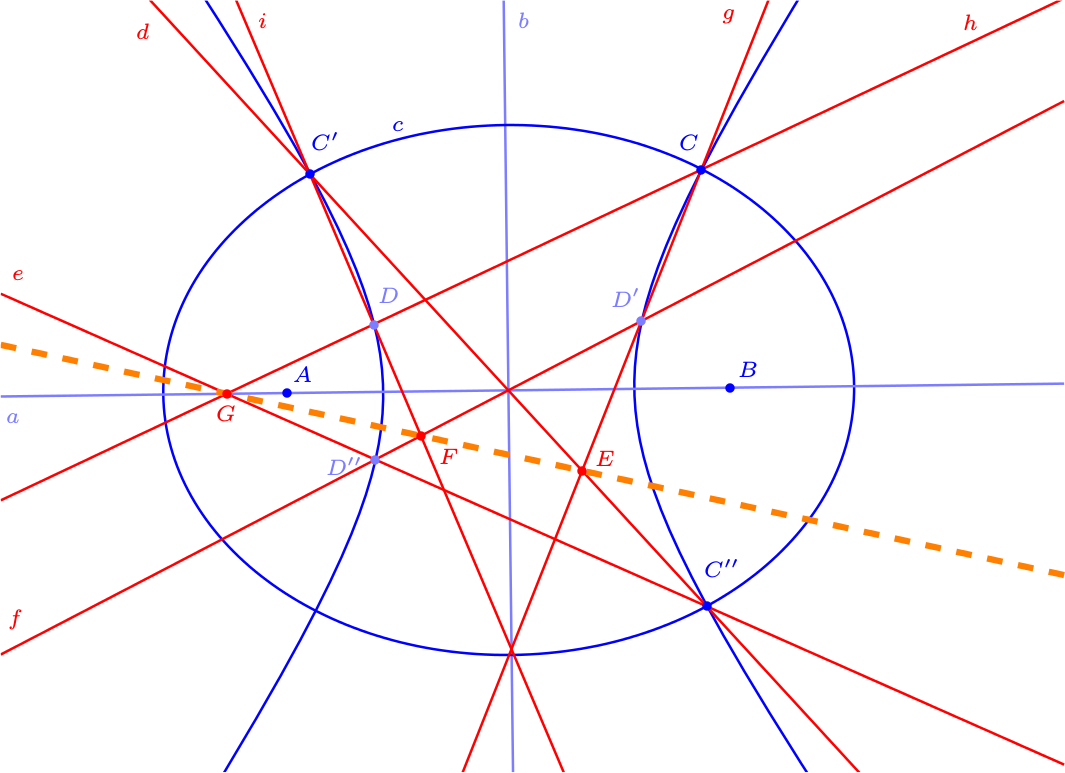}
\caption{A generalization of Pascal's hexagon theorem}
\label{Pascal}
\end{figure}

\begin{theorem}
Let $c$ be the union of an ellipse and a hyperbola, both defined with foci $A$, $B$ and circumpoint $C$. Let $a$ denote line $AB$ and let the perpendicular bisector of $a$ be $b$. Let $C'$ be the reflection of $C$ to the line $a$ and $C''$ to $b$. Also let us take an arbitrary point $D$ on $c$ and by reflection to $a$ and $b$, respectively, obtain points $D'$ and $D''$. Now the intersections of $CD$ and $C''D''$, $CD'$ and $C'C''$, moreover $C'D$ and $D'D''$ will be collinear.
\end{theorem}

A consequence of this example that some {\it formulas} can also be difficult to distinguish, and may require further investigation by using elimination and factorization with the help of a CAS.

In general, when a construction is given, it is important to identify geometrical {\it hypotheses} which are non-distinguishable from other geometrical hypotheses because they are translated with the same algebraic formula. When the prover {\it disproves} the respective statement in the algebraic translation, it should not be interpreted that the geometry statement was {\it false}. This is the case when attempting to prove that the internal bisectors of a triangle are concurrent: the algebraic translation actually {\it disproves} that the union of the internal and external angle bisectors are concurrent.

Also it is important to identify geometrical {\it theses} which are non-distinguishable from other geometrical theses because they are translated with the same algebraic formula. When the prover {\it proves} the respective statement in the algebraic translation, it should not be interpreted that the geometry statement was {\it true}.

\subsubsection{The implemented tools}

Apart from considering these issues, we managed to handle many typical classroom situations, and we report that most ``easy'' tools are implemented, and also some other tools from the ``middle'' and ``difficult to use'' toolset.

The following basic geometrical shapes are now implemented: segment, line, ray and vector, each defined by two points, circle defined with center and through point or through three points, angle, parabola with focus point and directrix, ellipse and hyperbola defined with two focus points. This table summarizes them, and also those tools which can operate on the basic geometrical shapes (the latter ones printed in italicized description, underlined objects are new enhancements compared to \cite{Kovacs_2015b}):

%\eject\vfill

%\begin{tabularx}{\textwidth}{c X X X}
\begin{tabularx}{\textwidth}{>{\hsize=.5\hsize}X >{\hsize=1.2\hsize}X >{\hsize=.5\hsize}X>{\hsize=1.8\hsize}X}

%\begin{table}[h!]
%\begin{tabularx}{\textwidth}{c X}
Tool & Description & Difficulty & Implementation remarks\\
\hline 
\parbox[c]{1cm}{\includegraphics[width=1cm]{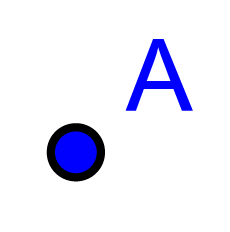}} & Point & easy &\\
\parbox[c]{1cm}{\includegraphics[width=1cm]{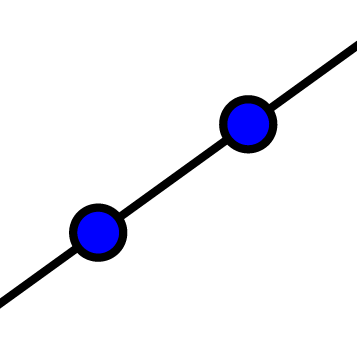}} & Line & easy &\\
\parbox[c]{1cm}{\includegraphics[width=1cm]{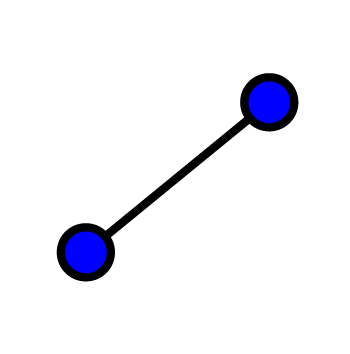}} & Segment & easy &\\
\parbox[c]{1cm}{\includegraphics[width=1cm]{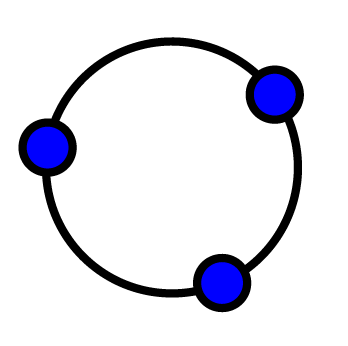}} & Circle through 3 Points & easy &\\
\parbox[c]{1cm}{\includegraphics[width=1cm]{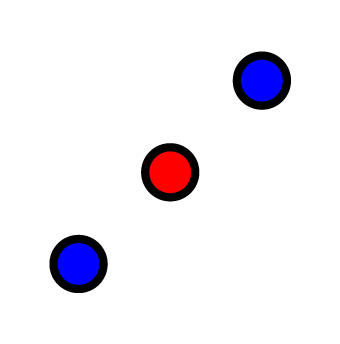}} & \textit{Midpoint or Center} & easy & points and segments\\
\parbox[c]{1cm}{\includegraphics[width=1cm]{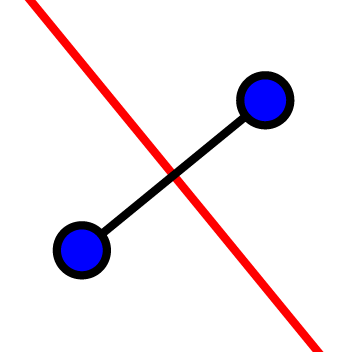}} & \textit{Perpendicular Bisector} & easy & at line and segment\\
\parbox[c]{1cm}{\includegraphics[width=1cm]{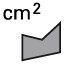}} & \ul{\textit{Area}} & easy & \ul{polygons of}\\
\parbox[c]{1cm}{\includegraphics[width=1cm]{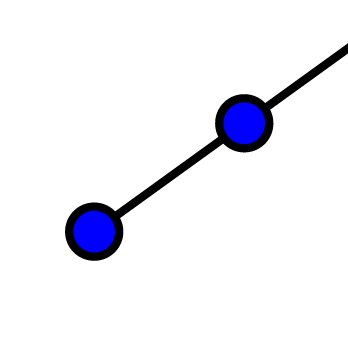}} & \ul{Ray} & middle &\\
\parbox[c]{1cm}{\includegraphics[width=1cm]{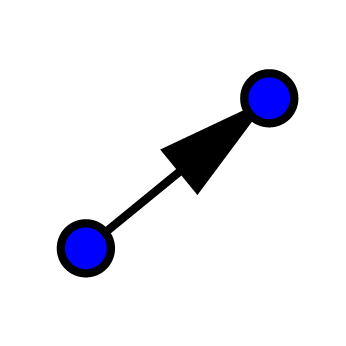}} & \ul{Vector} & middle &\\
\parbox[c]{1cm}{\includegraphics[width=1cm]{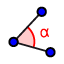}} & \ul{Angle} & middle &\\
\parbox[c]{1cm}{\includegraphics[width=1cm]{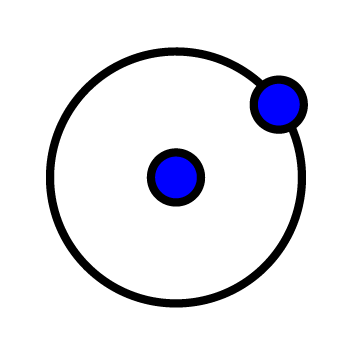}} & Circle with Center through Point & middle &\\
\parbox[c]{1cm}{\includegraphics[width=1cm]{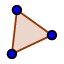}} & \ul{Polygon} & middle &\\
\parbox[c]{1cm}{\includegraphics[width=1cm]{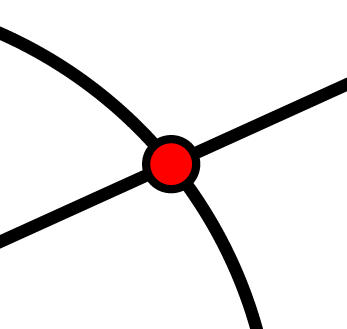}} & \textit{Intersect} & middle & line with line (cannot decide properly for segments), with circle, \ul{with parabola}, \ul{with ellipse}, \ul{with hyperbola}; circle with circle (for other conics we cannot decide properly)\\
\parbox[c]{1cm}{\includegraphics[width=1cm]{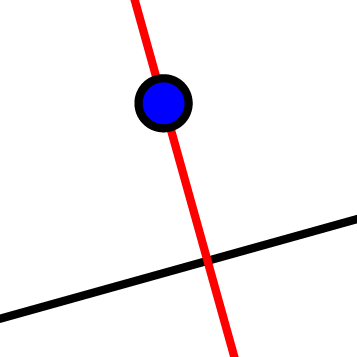}} & \textit{Perpendicular Line} & middle & at line through point\\
\parbox[c]{1cm}{\includegraphics[width=1cm]{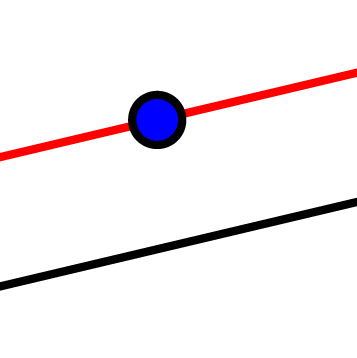}} & \textit{Parallel Line} & middle & with line through point\\
\parbox[c]{1cm}{\includegraphics[width=1cm]{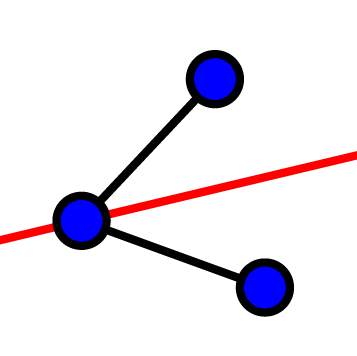}} & \ul{\textit{Angular Bisector}} & middle & \ul{angle defined by three points}\\
\parbox[c]{1cm}{\includegraphics[width=1cm]{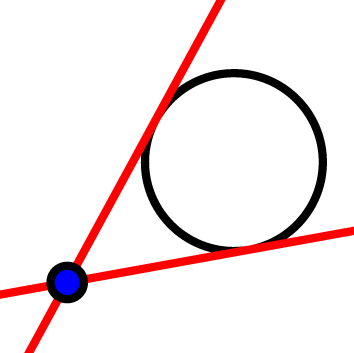}} & \ul{\textit{Tangents}} & middle & \ul{at circle, parabola, ellipse/hyperbola}\\
\parbox[c]{1cm}{\includegraphics[width=1cm]{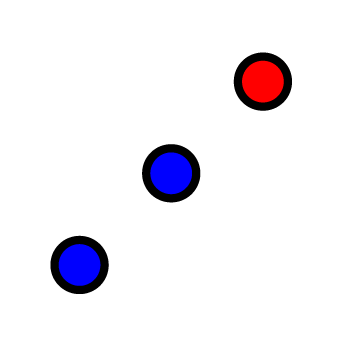}} & \ul{\textit{Reflect about Point}} & middle & \ul{point, line, circle, parabola, ellipse, hyperbola}\\
\parbox[c]{1cm}{\includegraphics[width=1cm]{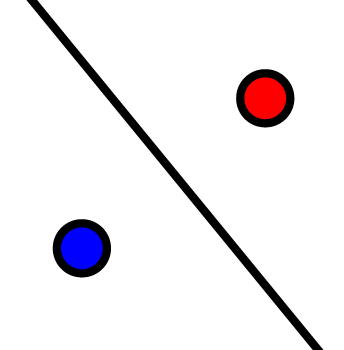}} & \ul{\textit{Reflect about Line}} & middle & \ul{point, line, circle, parabola}\\
\parbox[c]{1cm}{\includegraphics[width=1cm]{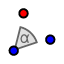}} & \ul{\textit{Rotate around Point}} & difficult & \ul{with known angles: $0\,^{\circ}$, $\pm 30\,^{\circ}$, $\pm 45\,^{\circ}$, $\pm 60\,^{\circ}$, $\pm 90\,^{\circ}$, $\pm 180\,^{\circ}$} \\
\parbox[c]{1cm}{\includegraphics[width=1cm]{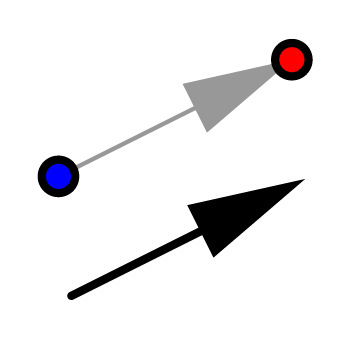}} & \ul{\textit{Translate by Vector}} & difficult  & \ul{point}\\
\parbox[c]{1cm}{\includegraphics[width=1cm]{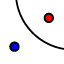}} & \ul{\textit{Reflect about Circle}} & --- & \ul{point, circle}\\
\parbox[c]{1cm}{\includegraphics[width=1cm]{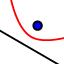}} & \ul{Parabola} & --- & \ul{with focus point and directrix}\\
\parbox[c]{1cm}{\includegraphics[width=1cm]{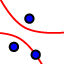}} & \ul{Hyperbola} & --- & \ul{with foci and circumpoint}\\
\parbox[c]{1cm}{\includegraphics[width=1cm]{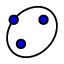}} & \ul{Ellipse} & --- & \ul{with foci and circumpoint}\\

\end{tabularx}
%\end{table}

The remaining, yet unimplemented ``easy'' tools in GeoGebra are: \textit{Conic through 5 Points} and \textit{Slope}. The former one is actually not widely used in the classroom, and the latter is a non-synthetic tool, that is, it is related to \textit{analytic geometry}. Some other missing, but planned features are listed in section \ref{future}.

\subsubsection{An example}

\begin{figure}
\includegraphics[width=\textwidth]{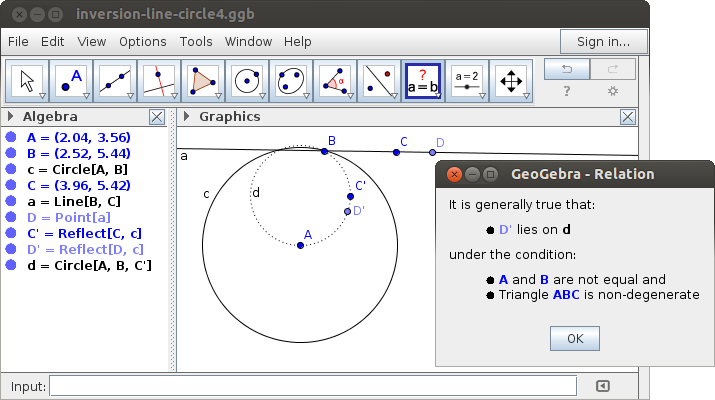}
\caption{Inversion translates lines to circles}
\label{gt-inv}
\end{figure}

\begin{theorem}Let $c$ be a circle with center $A$ and circumpoint $B$.
Let $a$ be a line through $B$ and $C$. Now---not considering some degenerate cases---reflecting line $a$ about $c$ the image is a circle, that is, for arbitrary point $D\in a$ its reflection $D'$ about $c$ always lies on the same circle (which is the circumcircle of points $A$, $B'(=B)$ and $C'$, where $B'$ and $C'$ are the mirror images of $B$ and $C$ about $c$, respectively).
\end{theorem}
In other words, {\it an inversion translates lines to circles in general}.
To use GeoGebra's Relation Tool (see Fig.~\ref{gt-inv}) one needs to set up the construction as described in the Algebra View on the left (either by selecting tools from the top, or by using commands in the Input Bar on the bottom). Finally one has to select the Relation Tool from the top and choose point $D'$ and line $d$ (or enter the command {\bf Relation[D,d]} in the Input Bar). GeoGebra now numerically checks if $D\in d$, the answer is yes, and the user can request a symbolical check by clicking on ``More$\ldots$''. Finally GeoGebra concludes that---under some {\it non-degeneracy} conditions---the statement is generally true.

From the computational point of view, GeoGebra here uses the Gr\"obner basis method. Thus it sets up the follwoing 6 equations in 13 variables, but point $A$ will be fixed to the origin (so there are only 11 variables remaining).
The following log information is printed only in debug mode in GeoGebra, including the timestamp in the first column:

{\scriptsize
\begin{lstlisting}[language=mylog]
19:48:26.550 // Free point A(v1,v2)
19:48:26.550 // Free point B(v3,v4)
19:48:26.550 c = Circle[A, B] /* Circle through B with center A */
19:48:26.551 // Free point C(v5,v6)
19:48:26.551 a = Line[B, C] /* Line through B, C */
19:48:26.551 D = Point[a] /* Point on a */
19:48:26.555 // Constrained point D(v7,v8)
19:48:26.555 Hypotheses:
19:48:26.555 1. -1*v7*v6+v8*v5+v7*v4+-1*v5*v4+-1*v8*v3+v6*v3
19:48:26.556 C' = Mirror[C, c] /* C mirrored at c */
19:48:26.560 // Constrained point C'(v9,v10)
19:48:26.561 2. -1*v9*v6^2+-1*v9*v5^2+v5*v4^2+v5*v3^2+2*v9*v6*v2+-2*v5*v4*v2+-1*v9*v2^2+v5*v2^2+v6^2*v1+2*v9*v5*v1+v5^2*v1+-1*v4^2*v1+-2*v5*v3*v1+-1*v3^2*v1+-2*v6*v2*v1+2*v4*v2*v1+-1*v9*v1^2+-1*v5*v1^2+2*v3*v1^2
19:48:26.562 3. -1*v10*v6^2+-1*v10*v5^2+v6*v4^2+v6*v3^2+2*v10*v6*v2+v6^2*v2+v5^2*v2+-2*v6*v4*v2+-1*v4^2*v2+-1*v3^2*v2+-1*v10*v2^2+-1*v6*v2^2+2*v4*v2^2+2*v10*v5*v1+-2*v6*v3*v1+-2*v5*v2*v1+2*v3*v2*v1+-1*v10*v1^2+v6*v1^2
19:48:26.562 D' = Mirror[D, c] /* D mirrored at c */
19:48:26.566 // Constrained point D'(v11,v12)
19:48:26.567 4. -1*v11*v8^2+-1*v11*v7^2+v7*v4^2+v7*v3^2+2*v11*v8*v2+-2*v7*v4*v2+-1*v11*v2^2+v7*v2^2+v8^2*v1+2*v11*v7*v1+v7^2*v1+-1*v4^2*v1+-2*v7*v3*v1+-1*v3^2*v1+-2*v8*v2*v1+2*v4*v2*v1+-1*v11*v1^2+-1*v7*v1^2+2*v3*v1^2
19:48:26.568 5. -1*v12*v8^2+-1*v12*v7^2+v8*v4^2+v8*v3^2+2*v12*v8*v2+v8^2*v2+v7^2*v2+-2*v8*v4*v2+-1*v4^2*v2+-1*v3^2*v2+-1*v12*v2^2+-1*v8*v2^2+2*v4*v2^2+2*v12*v7*v1+-2*v8*v3*v1+-2*v7*v2*v1+2*v3*v2*v1+-1*v12*v1^2+v8*v1^2
19:48:26.568 Hypotheses have been processed.
19:48:26.574 substitutions: {v1=0, v2=0}
19:48:26.574 Thesis reductio ad absurdum (denied statement)...
19:48:26.586 6. -1+-1*v13*v11*v10^2*v4+v13*v12^2*v9*v4+v13*v11^2*v9*v4+-1*v13*v11*v9^2*v4+v13*v11*v10*v4^2+-1*v13*v12*v9*v4^2+-1*v13*v12^2*v10*v3+-1*v13*v11^2*v10*v3+v13*v12*v10^2*v3+v13*v12*v9^2*v3+v13*v11*v10*v3^2+-1*v13*v12*v9*v3^2+v13*v11*v10^2*v2+-1*v13*v12^2*v9*v2+-1*v13*v11^2*v9*v2+v13*v11*v9^2*v2+-1*v13*v11*v4^2*v2+v13*v9*v4^2*v2+v13*v12^2*v3*v2+v13*v11^2*v3*v2+-1*v13*v10^2*v3*v2+-1*v13*v9^2*v3*v2+-1*v13*v11*v3^2*v2+v13*v9*v3^2*v2+-1*v13*v11*v10*v2^2+v13*v12*v9*v2^2+v13*v11*v4*v2^2+-1*v13*v9*v4*v2^2+-1*v13*v12*v3*v2^2+v13*v10*v3*v2^2+v13*v12^2*v10*v1+v13*v11^2*v10*v1+-1*v13*v12*v10^2*v1+-1*v13*v12*v9^2*v1+-1*v13*v12^2*v4*v1+-1*v13*v11^2*v4*v1+v13*v10^2*v4*v1+v13*v9^2*v4*v1+v13*v12*v4^2*v1+-1*v13*v10*v4^2*v1+v13*v12*v3^2*v1+-1*v13*v10*v3^2*v1+-1*v13*v11*v10*v1^2+v13*v12*v9*v1^2+v13*v11*v4*v1^2+-1*v13*v9*v4*v1^2+-1*v13*v12*v3*v1^2+v13*v10*v3*v1^2
19:48:26.592 Eliminating system in 11 variables (6 dependent)
\end{lstlisting}
}

Then the underlying CAS (here Giac) eliminates variables {\tt v8}, {\tt v9}, {\tt v10}, {\tt v11}, {\tt v12} and {v13} to describe non-degeneracy conditions between the coordinates of the free points. The obtained equation system in factorized form is produced in the following output (which is compatible with Singular's arrays, cf. \cite[p.~146]{Kovacs_2015b}):

{\scriptsize
\begin{lstlisting}
[1]:
 [1]:
  _[1]=1
  _[2]=-v6^2-v5^2
  _[3]=v7
 [2]: 1,1,1
[2]:
 [1]:
  _[1]=1
  _[2]=v4^2+v3^2
  _[3]=v5
  _[4]=v7
 [2]: 1,1,1,1
...
[12]:
 [1]:
  _[1]=1
  _[2]=v4*v6*v7^3-v4*v6*v7^2*v3-v4*v6*v7*v5^2+v4*v6*v5^2*v3-v6^2*v5*v3^2+v7^3*v5*v3-v7^2*v5*v3^2-v7*v5^3*v3
 [2]: 1,1
[13]:
 [1]:
  _[1]=1
  _[2]=v4^2+v3^2
  _[3]=-v5*v4+v6*v3
  _[4]=-1
 [2]: 2,2,1,1
\end{lstlisting}
} % \scriptsize

This is interpreted by GeoGebra as 13 possible sets of degeneracy conditions. Here---because of its geometrical meaning, simplicity and being fully synthetic---the 13th set will be selected, which means: ``if \verb|(v4^2+v3^2)^2*(-v5*v4+v6*v3)| differs from $0$, then the thesis will be true on all possible values of the coordinates of the free points''. Since $A=(0,0)$, $B={\tt(v3,v4)}$ and $C={\tt(v5,v6)}$, this clearly means that the two non-degeneracy conditions being shown are ``$A$ differs from $B$'' (that is, circle $c$ is non-degenerate) and ``$A$, $B$ and $C$ are not collinear'' (that is, such a line must be chosen for $a$ which is not going through the center of $c$).

Finally, GeoGebra concludes that

{\scriptsize
\begin{lstlisting}
19:48:26.714 Statement is GENERALLY TRUE
19:48:26.714 Benchmarking: 487 ms
19:48:26.717 OUTPUT for ProveDetails: null = {true, {"AreCollinear[A,B,C]", "AreEqual[A,B]"}}
\end{lstlisting}
}

This computation is done faster than half of a second.\footnote{The steps and the output for computing this example have been simplified to fit this paper. See also \cite{Kovacs_2015b} for the detailed algorithm of symbolical checking in the Relation Tool.}

\subsubsection{Technical notes}

Technically speaking, GeoGebra is a Java application. From the developer's point of view, the Java public interface SymbolicParametersBotanaAlgo has to be implemented in GeoGebra's Algo* classes by creating suitable algebraic equations (and corresponding new variables) to describe the symbolic background of a newly used tool.\footnote{See
\url{https://dev.geogebra.org/javadoc/common/org/geogebra/common/kernel/algos/SymbolicParametersBotanaAlgo.html}
for a recent list of the implemented classes.}

To check the validity of a thesis, the public interface SymbolicParametersBotanaAlgoAre must be implemented.\footnote{See
\url{https://dev.geogebra.org/javadoc/common/org/geogebra/common/kernel/algos/SymbolicParametersBotanaAlgoAre.html} for a recent list.} Currently the following checks are implemented: collinearity, concurrency, concyclicity, congruency, equality, parallelism, perpendicularity, incidence, and formula checking (to prove equations).

\subsection{Benchmarks}
\label{benchmarks}

In our improvements the benchmark suite was extended by additional 140 theorems. 57 of these extra tests were chosen from \cite{Chou_1987}---these tests were computed in Chou's book by using Wu's \cite{Wu_1984} characteristic method.

Here we summarize our results by sharing a list of the recent benchmarking outputs. GeoGebra's prover benchmarking system is available as a command line tool in its source folder \texttt{test/scripts/benchmark/prover/}.

\subsubsection{Desktop version}
GeoGebra's desktop version runs as a Java native application on the mostly used operating system platforms including Windows, Mac OS X and Linux. Due to the internally used native Giac CAS each platform requires its own compiled version of the embedded computer algebra system.

The following table is the output of the "jar-paper" scenario, launched by the command line \texttt{xvfb-run ./runtests -S jar-paper -r} in this folder. This scenario tests the {\bf Prove} command exclusively. See also \cite{Botana_2015}.

\begin{itemize}
\item The first column abbreviates the name of the test cases.
\item Column E1 ("Engine 1") refers to Recio's exact check method programmed by Simon Weitzhofer.
\item Column E2 ("Engine 2") refers to the Gr\"obner basis method via SingularWS (also known as Botana's method) programmed by the authors of this paper. (See \cite{Botana_2014} for more on SingularWS.)
\item Column E2/Giac refers to Gr\"obner basis method via the Giac computer algebra tool (instead of SingularWS) programmed by Bernard Parisse and the authors of this paper.
\item Column E3a ("Engine 3a") refers to OpenGeoProver's Wu's method implementation programmed by
Ivan Petrovi\'c and Predrag Jani\v{c}i\'c.
\item Column E3b ("Engine 3b") refers to OpenGeoProver's Area method programmed by Damien Desfontaines.
\item The Auto approach refers to the automatic selection of methods which is already
 implemented in GeoGebra and it usually starts with "Engine 1" and
 then it continues with "Engine 2" (either via SingularWS or
 Giac: if SingularWS is available, then in SingularWS, otherwise in Giac). If
 the Gr\"obner basis method is not conclusive, then "Engine 3a" is tried.
 If it is not conclusive either, then OpenGeoProver's
 Area method (Engine 3b) is used.
\end{itemize}

See \cite{Botana_2015} for more details about the used methods.
Explanation of the used colors:

\begin{itemize}
\item Green means that the test returns a correct yes/no answer. Intensity of green
means speed (the lighter the slower). Numbers are in milliseconds.
\item Pink means that GeoGebra returns the wrong answer.
\item Yellow means the output is not conclusive, thus using this method
GeoGebra shows "undefined", i.e. there is no error here.
\item The R. ("Result") column provides some extra information about the result, such as
f ("false") when the statement was false on purpose.
\end{itemize}	
The S. ("Speed") column shows the timing. Highlighted entries are the best results, italicized entries are the slowest (but working) results in a row. The test cases are also available for download in GeoGebra's .ggb format from the GeoGebra online source code directly.

For testing we used a PC with 16 GB RAM, 8 $\times$ Intel(R) Core(TM) i7 CPU 860 @ 2.80GHz, and Linux Mint 17.2.

{
\tiny{
\begin{longtable}{|l|*{6}{cr|}}
\hline
{\multirow{2}{*}{\bf Test}}
& \multicolumn{2}{c|}{\bf E1}
& \multicolumn{2}{c|}{\bf E2}
& \multicolumn{2}{c|}{\bf E2/Giac}
& \multicolumn{2}{c|}{\bf E3a}
& \multicolumn{2}{c|}{\bf E3b}
& \multicolumn{2}{c|}{\bf Auto}
\\
&R.&S.
&R.&S.
&R.&S.
&R.&S.
&R.&S.
&R.&S.
\\ \hline
\cellcolor{blue!10}lines-parallel 
& \cellcolor{green!60}f & \cellcolor{green!60}{\bf 3} 
& \cellcolor{green!50}f & \cellcolor{green!50}{ 35} 
& \cellcolor{yellow!25} & \cellcolor{yellow!25}{ 185} 
& \cellcolor{green!40}f & \cellcolor{green!40}{ 65} 
& \cellcolor{green!40}f & \cellcolor{green!40}{\sl 69} 
& \cellcolor{green!60}f & \cellcolor{green!60}{ 7} 
\\ \hline
\cellcolor{blue!10}midpoint-third 
& \cellcolor{yellow!25} & \cellcolor{yellow!25}{ 3} 
& \cellcolor{green!30}f & \cellcolor{green!30}{\bf 217} 
& \cellcolor{yellow!25} & \cellcolor{yellow!25}{ 201} 
& \cellcolor{yellow!25} & \cellcolor{yellow!25}{ 58} 
& \cellcolor{yellow!25} & \cellcolor{yellow!25}{ 72} 
& \cellcolor{yellow!25} & \cellcolor{yellow!25}{ 311} 
\\ \hline
\cellcolor{blue!10}orthocente$\ldots$ar 
& \cellcolor{yellow!25} & \cellcolor{yellow!25}{ 4} 
& \cellcolor{green!50}f & \cellcolor{green!50}{\bf 48} 
& \cellcolor{yellow!25} & \cellcolor{yellow!25}{ 202} 
& \cellcolor{yellow!25} & \cellcolor{yellow!25}{ 75} 
& \cellcolor{yellow!25} & \cellcolor{yellow!25}{ 85} 
& \cellcolor{yellow!25} & \cellcolor{yellow!25}{ 312} 
\\ \hline
\cellcolor{blue!10}points-col$\ldots$ar 
& \cellcolor{green!60}f & \cellcolor{green!60}{\bf 3} 
& \cellcolor{green!50}f & \cellcolor{green!50}{ 34} 
& \cellcolor{yellow!25} & \cellcolor{yellow!25}{ 196} 
& \cellcolor{green!40}f & \cellcolor{green!40}{ 62} 
& \cellcolor{green!40}f & \cellcolor{green!40}{\sl 64} 
& \cellcolor{green!60}f & \cellcolor{green!60}{ 3} 
\\
\hline
\end{longtable}
}
}

We highlight that:
\begin{itemize}
\item Our theorem corpus has a significant number of test cases. Cf. \cite{Jani_i__2007}.
\item The best performing theorem prover---when using our corpus---is the complex algebraic geometry prover via Singular \cite{Singular}. Here the \cite[chapter 6, \S 4]{Cox_2007} algorithm was used. Timing is remarkably under one second in most test cases.
\item The table can be misleading when investigating other columns. Actually, there is no implementation for intersections with conics in GeoGebra for Recio's method. Also E2/Giac can use a different algorithm with better (but slightly slower) results. Some GeoGebra commands are not yet implemented in the communication layer between GeoGebra and OpenGeoProver, that is, columns E3a and E3b show only a limited amount of positive test cases.
\item For the end user the significant case is the last column, since Singular is disabled by default to ensure the same behavior on offline and online runs.
\end{itemize}

\subsubsection{Web version}
The web version runs in a web browser. All major browsers including Google Chrome, Mozilla Firefox and Internet Explorer are supported.

The following table was generated by using the command line \verb|xvfb-run ./runtest -p "Auto Web" -r| in this folder. It compares the outputs of the {\bf Prove} command in the desktop version ("Auto") and the web version ("Web").

{
\tiny{
\begin{longtable}{|l|*{2}{cr|}}
\hline
{\multirow{2}{*}{\bf Test}}
& \multicolumn{2}{c|}{\bf Auto}
& \multicolumn{2}{c|}{\bf Web}
\\
&R.&S.
&R.&S.
\\ \hline
\cellcolor{blue!10}lines-parallel 
& \cellcolor{green!60}f & \cellcolor{green!60}{ 7} 
& \cellcolor{green!60}f & \cellcolor{green!60}{\bf 11} 
\\ \hline
\cellcolor{blue!10}midpoint-third 
& \cellcolor{yellow!25} & \cellcolor{yellow!25}{ 297} 
& \cellcolor{yellow!25} & \cellcolor{yellow!25}{ 779} 
\\ \hline
\cellcolor{blue!10}orthocente$\ldots$ar 
& \cellcolor{yellow!25} & \cellcolor{yellow!25}{ 303} 
& \cellcolor{yellow!25} & \cellcolor{yellow!25}{ 626} 
\\ \hline
\cellcolor{blue!10}points-col$\ldots$ar 
& \cellcolor{green!60}f & \cellcolor{green!60}{ 5} 
& \cellcolor{green!60}f & \cellcolor{green!60}{\bf 9} 
\\ \hline
\cellcolor{blue!10}points-equal 
& \cellcolor{green!60}f & \cellcolor{green!60}{ 4} 
& \cellcolor{green!60}f & \cellcolor{green!60}{\bf 11} 
\\ \hline
\cellcolor{blue!10}Pythagoras$\ldots$n4 
& \cellcolor{red!25}t & \cellcolor{red!25}{ 311} 
& \cellcolor{yellow!25} & \cellcolor{yellow!25}{ 612} 
\\
\hline
\end{longtable}
}}

We highlight that:
\begin{itemize}
\item The web version does not return any incorrect output in any cases.
\item It is properly working in 125/200 cases (62.5\%) which is 86.2\% of the performance ratio of the desktop version.
\item The web version is definitely slower than the desktop version by a ratio between 2 and 6.
\item Despite its limited availability and speed, the web version is already applicable in many classroom situations. The users only need a web browser which should be accessed not only on desktop computers and laptops, but also on tablets and mobile phones.
\end{itemize}

\subsubsection{Theorems in the classrooms}
To sum up, we list some important theorems which are usually discussed in secondary schools. Now they can be proven with GeoGebra's help, that is, at least a yes/no answer is provided for many theorems, including:
\begin{itemize}
\item \ul{The Pythagorean theorem}. \ul{The intercept theorem}. \ul{The geometric mean and cathetus theorems}. Thales' theorem.
\item Concurrency of medians, bisectors, altitudes. Euler line. The midline theorem, Varignon's theorem. The nine points circle. Simson's theorem.
\item \ul{The angle bisector theorem}.
\item Basic properties of translations and \ul{rotations}.
\item \ul{Basic properties of reflections about a point, a line or a circle.}
\item \ul{Ceva's theorem}, \ul{Menelaus' theorem}. Desargues's theorem, Pappus' theorem.
\item \ul{Basic properties of conic sections (including tangents).}
\end{itemize}
The underlined theorems can be proven with the internal complex algebraic geometry prover in GeoGebra by using the enhancements implemented in the last milestone in our work.

\section{Future work}
\label{future}

Finally, we summarize the currently planned new features in the forthcoming versions of GeoGebra.

\begin{tabularx}{\textwidth}{c X X}

%\begin{table}[h!]
%\begin{tabularx}{\textwidth}{c X}
GeoGebra tool & Description & To implement\\
\hline 
\parbox[c]{1cm}{\includegraphics[width=1cm]{mode_area.png}} & \textit{Area} & of conics\\
\parbox[c]{1cm}{\includegraphics[width=1cm]{mode_translatebyvector.png}} & \textit{Translate by Vector} & line, segment, ray, circle, parabola, ellipse, hyperbola, polygons\\
\parbox[c]{1cm}{\includegraphics[width=1cm]{mode_mirroratline.png}} & \textit{Reflect about Line} & ellipse, hyperbola\\
\parbox[c]{1cm}{\includegraphics[width=1cm]{mode_mirroratcircle.png}} & \textit{Reflect about Circle} & line\\
\parbox[c]{1cm}{\includegraphics[width=1cm]{mode_rotatebyangle.png}} & \textit{Rotate around Point} & general angles \\

\end{tabularx}
%\eject\vfill

There is still room for further enhancements:

\begin{itemize}
\item Improve formula handling by eliminating non-squared quantities automatically and identifying formulas for a correct decision about the truth of the statement.
\item Currently it is not possible to mirror a line about a circle directly: in this case the implementation should handle that the object type is changing from line to circle in general.
\item The {\bf ShowProof} command \cite{Kovacs_2014} might be implemented in cases when a readable proof can be produced automatically.
\item Allow proofs for 3D Euclidean geometry (cf. \cite{Roanes_2007}).
\item Improve Gr\"obner bases computations in Giac to implement transcendent coefficients (see \cite[chapter 6, \S 4]{Cox_2007}). This would speed up computations in a number of cases which are currently infeasible: an indirect reduction of variables would be achieved in this way.
\item GeoGebra's {\bf LocusEquation} command is capable of computing algebraic loci \cite{Botana_2014}. It would be possible to unify the code base for the locus and the prover subsystem, and the unified system could be maintained and improved easier.
\end{itemize}
Also implementing conic sections for Recio's exact check method would speed up GeoGebra's proofs significantly.

\section*{Acknowledgments}

The theorem proving subsystem in GeoGebra is a joint work with contributions from several researchers, programmers and teachers. We are especially thankful to Bernard Parisse for improving the Giac CAS to be competitive to Singular and some commercial systems, making possible that GeoGebra has a robust embedded theorem prover also.

We are grateful to Tom\'as Recio, Predrag Jani\v{c}i\'c, Julien Narboux and Francisco Botana for their useful hints to improve the text of this paper, and to Markus Hohenwarter and Judit Robu for
supporting our work.

%\bibliography
%{bibliography/converted_to_latex.bib}

\end{document}